\documentclass[runningheads]{llncs}
\usepackage[T1]{fontenc}
\usepackage{graphicx,verbatim}
\usepackage{amsmath}
\usepackage{amssymb}
\usepackage{booktabs}

\usepackage{url}            
\usepackage{booktabs}       
\usepackage{amsfonts}       
\usepackage{nicefrac}       
\usepackage{microtype}      
\usepackage{xcolor}         
\usepackage{amsmath}
\usepackage{afterpage}

\usepackage{times}  
\usepackage{helvet}  
\usepackage{courier}  

\usepackage{caption} 
\usepackage{algorithm}
\usepackage{algorithmic}

\usepackage{newfloat}
\usepackage{listings}
\usepackage{multirow}

\usepackage{tabularray}
\usepackage[capitalize]{cleveref}
\crefname{section}{Sec.}{Secs.}
\Crefname{section}{Section}{Sections}
\Crefname{table}{Table}{Tables}
\crefname{table}{Tab.}{Tabs.}

\begin{document}
\title{Pre-trained Models Succeed in Medical Imaging with Representation Similarity Degradation}
%

\author{
\textbf{Wenqiang Zu}$^{1,4}$, 
\textbf{Shenghao Xie}$^{2}$, 
\textbf{Hao Chen}$^{3}$, 
\textbf{Lei Ma}$^{2,4}$\thanks{Corresponding author.}
}

\authorrunning{W. Zu et al.}
\institute{Institute of Automation, Chinese Academy of Sciences \and
Academy for Advanced Interdisciplinary Studies, National Biomedical Imaging Center, National Key Laboratory for Multimedia Information Processing, Peking University\and
School of Chemical Sciences, University of Chinese Academy of Sciences \and 
Beijing Academy of Artificial Intelligence \\
    \email{zuwenqiang2022@ia.ac.cn, shenghaoxie@stu.pku.edu.cn, lei.ma@pku.edu.cn}}

\maketitle              
\begin{abstract}
This paper investigates the critical problem of representation similarity evolution during cross-domain transfer learning, with particular focus on understanding why pre-trained models maintain effectiveness when adapted to medical imaging tasks despite significant domain gaps. The study establishes a rigorous problem definition centered on quantifying and analyzing representation similarity trajectories throughout the fine-tuning process, while carefully delineating the scope to encompass both medical image analysis and broader cross-domain adaptation scenarios. Our empirical findings reveal three critical discoveries: the potential existence of high-performance models that preserve both task accuracy and representation similarity to their pre-trained origins; a robust linear correlation between layer-wise similarity metrics and representation quality indicators; and distinct adaptation patterns that differentiate supervised versus self-supervised pre-training paradigms. The proposed similarity space framework not only provides mechanistic insights into knowledge transfer dynamics but also raises fundamental questions about optimal utilization of pre-trained models. These results advance our understanding of neural network adaptation processes while offering practical implications for transfer learning strategies that extend beyond medical imaging applications. The code will be available once accepted.


\end{abstract}

\section{Introduction}
The rapid development of vision foundation models, driven by advancements in large-scale datasets~\cite{deng2009imagenet,schuhmann2022laion} and pre-training paradigms~\cite{he2022masked,liu2021self,he2020momentum}, has yielded powerful pre-trained architectures~\cite{dosovitskiy2020image,kirillov2023segment,radford2021learning,oquab2023dinov2} with remarkable cross-domain generalization capabilities. These models demonstrate unprecedented adaptability to downstream tasks, even under substantial domain shifts and limited training data, making their application to specialized domains like medical image analysis~\cite{kim2022transfer,dutt2023parameter,wu2023medical,ma2024segment,huix2024natural,matsoukas2023pretrained} both practical and cost-effective. However, current methodologies predominantly treat pre-trained models as black-box initialization tools, lacking mechanistic analysis of how and why their representations remain effective despite significant domain discrepancies—a critical oversight given the growing prevalence of model-task mismatches~\cite{xie2024towards}. This gap in understanding representation similarity dynamics during adaptation forms the central focus of our investigation.

\begin{figure}[!t]
    \captionsetup{singlelinecheck=off, justification=raggedright,labelsep=period}
  \includegraphics[width=1\textwidth]{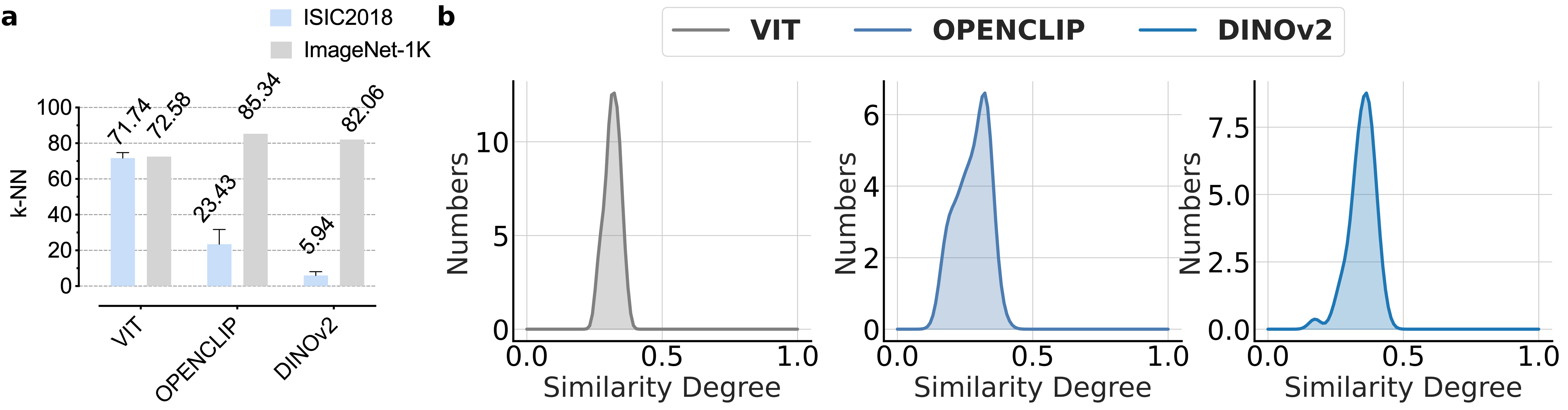}
  \caption{(a) The k-NN evaluation results on ImageNet-1K between fine-tuned models (fine-tuned on ISIC2018, blue bar) and pre-trained models (gray bar). (b) Similarity distribution curves for models fine-tuned on ISIC2018 dataset~\cite{codella2019skin} (20 runs/method), recording final representation similarity with pre-trained models.}
  \label{fig:fig1}
\end{figure}

Our work addresses two fundamental questions: (1) how pre-trained knowledge evolves during cross-domain transfer, and (2) what metrics best capture representation stability under domain shifts. Through systematic analysis of similarity trajectories in both parameter space and feature representations, we establish a rigorous framework for evaluating model adaptation that extends beyond medical imaging to general cross-domain scenarios. Empirical evidence from our k-NN evaluation (Fig.~\ref{fig:fig1}a) reveals a striking divergence: unsupervised models like DINOv2~\cite{oquab2023dinov2} and OPENCLIP~\cite{radford2021learning} exhibit significantly reduced post-adaptation performance compared to supervised counterparts, despite their theoretical versatility. This observation challenges conventional assumptions about pre-training superiority~\cite{bommasani2021opportunities} and motivates our core hypothesis—that representation similarity preservation, rather than mere initialization quality, governs effective knowledge transfer. The critical unresolved question emerges: does unsupervised pre-training inherently limit adaptation capacity, or does supervised fine-tuning induce representational instability in these models?

The conventional adaptation paradigm rests on a fundamental premise: that features encoded through large-scale pre-training inherently align with latent data distributions across domains~\cite{huh2024platonic,yao2024unifying}. While recent parameter-efficient tuning methods like EPT~\cite{zu2024embedded} demonstrate effectiveness in medical imaging through distribution calibration, they leave unresolved critical questions about similarity evaluation between pre-trained and adapted models. This assumption underpins the expectation that fine-tuned models should maintain representational continuity with their pre-trained counterparts when bridging domain gaps between natural and medical imaging. Our investigation challenges this orthodoxy through systematic similarity analysis~\cite{klabunde2023similarity}, revealing a critical discrepancy between theoretical assumptions and empirical observations in cross-domain adaptation.

We investigate fine-tuning principles through three core aspects. Comparative analysis of supervised versus unsupervised pre-trained models via ImageNet-1K k-NN classification reveals unsupervised models exhibit enhanced semantic learning during supervised fine-tuning (evidenced by attention visualization) yet unexpected robustness fragility. Systematic measurement of representation similarity using established metrics~\cite{klabunde2023similarity} (Fig. 1(b)) uncovers a paradox where high-performing fine-tuned models show significantly reduced similarity, raising crucial questions about knowledge retention~\cite{ramasesh2021effect}. Visual and quantitative analyses reveal consistent correlations between representation similarity patterns and pre-trained models' intrinsic metrics, leading to our proposal of \emph{similarity space} theory - identifying optimal subspaces where models maintain accuracy while preserving pre-training advantages. Practical applications demonstrate that pre-training metrics strongly predict post-fine-tuning characteristics, with experimental validation showing key indicators like calibration error~\cite{wang2023calibration} and k-NN evaluations can be projected from pre-trained representations through similarity analysis. 

This contribution advances fine-tuning theory and practice, offering new insights into knowledge transfer dynamics while enabling model selection and performance prediction methodologies.

\begin{figure*}[!t]
\captionsetup{singlelinecheck=off, justification=raggedright,labelsep=period}
  \includegraphics[width=1\textwidth]{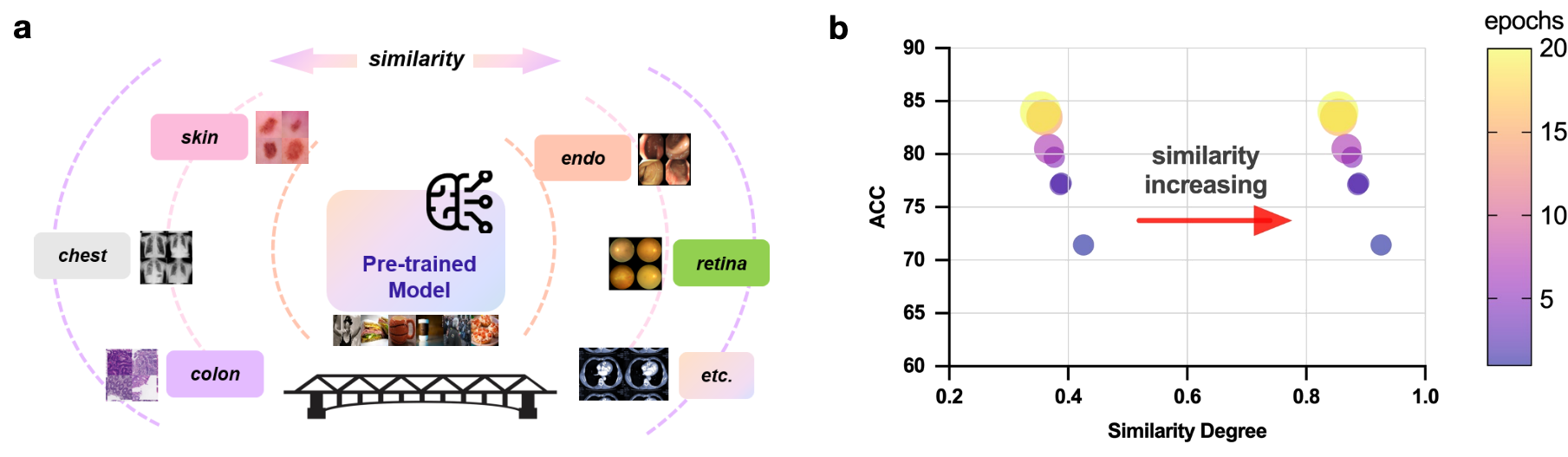}
  \caption{Similarity space: (a) Fine-tuned models show varied similarity levels in pre-trained parameter space. (b) Models within the similarity space demonstrate both performance and similarity.}
  \label{fig:fig2}
\end{figure*}
\section{Method}

\subsection{Pre-trained Models}
We select three pre-trained models: the supervised ViT~\cite{dosovitskiy2020image} model (VIT), the image-text contrastive learning-based OPENCLIP, and the unsupervised DINOv2. This comparison helps analyze fine-tuning differences between supervised and unsupervised training while maintaining structural consistency with ViT-Base. Supervised ViT model (VIT) is pre-trained on ImageNet-1K, while OPENCLIP model aligns image-text representations using contrastive learning on LAION-2B~\cite{schuhmann2022laion}, and DINOv2 model follows the DINO~\cite{zhang2022dino} framework for unsupervised learning on a larger dataset.

\subsection{Datasets}  
The MedFM datasets~\cite{wang2023real}, ISIC2018~\cite{codella2019skin} dataset and APTOS2019~\cite{aptos2019} dataset are used for fine-tuning. These datasets cover various modalities, organs, and regions with comparable data scales. We select them based on classification task diversity, including multiclass, binary, and multi-label classification. ISIC2018 dataset consists of seven skin lesion categories with 10,015 training and 1,512 test images. APTOS2019 dataset is designed for diabetic diagnosis through fundus images, including five severity grades with 2,930 training and 366 test images. MedFM-Chest involves 19 chest diagnosis categories with 2,140 training and 3,869 test images, evaluated using mAP. MedFM-Colon focuses on gastrointestinal lesion detection with 5,654 training images and 7,654 test images. MedFM-Endo includes four categories with 1,810 training and 2,936 test images, also evaluated using mAP.

\subsection{{Evaluation Metrics}}  
Centered Kernel Alignment (CKA)~\cite{kornblith2019similarity} measures similarity between two representations. For kernel matrices $\mathbf{K}, \mathbf{L} \in \mathbb{R}^{n \times n}$, CKA is defined as  

\begin{equation}  
\operatorname{CKA}(\mathbf{K}, \mathbf{L}) = \frac{\operatorname{tr}(\mathbf{K}\,H\,\mathbf{L}\,H)}{\sqrt{\operatorname{tr}(\mathbf{K}\,H\,\mathbf{K}\,H) \operatorname{tr}(\mathbf{L}\,H\,\mathbf{L}\,H)}}  
\end{equation}  
where $H = I_n - \frac{1}{n} \mathbf{1}\mathbf{1}^{\mathrm{T}}$ is the centering matrix, $I_n$ is the identity matrix, and $\mathbf{1} \in \mathbb{R}^{n}$ is a vector of ones.

The Expected Calibration Error (ECE)~\cite{wang2023calibration} quantifies the discrepancy between a classifier's accuracy and predicted confidence. For $n$ samples grouped into $K$ bins $\{\beta_k\}_{k=1}^K$, ECE is defined as  

\begin{equation}  
\text{ECE} = \frac{1}{n} \sum_{k=1}^{K} |\beta_k| \cdot \left| \text{Acc}(\beta_k) - \text{Conf}(\beta_k) \right|  
\end{equation}  
where $\text{Acc}(\beta_k)$ and $\text{Conf}(\beta_k)$ are the accuracy and confidence of bin $\beta_k$.

\subsection{Formal Definition of Similarity Space}
The characterization of \emph{similarity space}, as shown in Fig.~\ref{fig:fig2}, is established through rigorous operator-theoretic constructs governing parameterized model behaviors. Let 
$
\phi: \Theta \times \mathcal{X} \to \mathcal{R}
$
denote the representation mapping and 
$
\psi: \Theta \times \mathcal{X} \to \mathcal{Y}
$
the prediction mapping, where $\Theta$ parametrizes the model space, $\mathcal{X}$ is the input domain, $\mathcal{R}$ the representation manifold, and $\mathcal{Y}$ the label space. For a pretrained model $\theta^* \in \Theta$ and its fine-tuned variant $\theta \in \Theta$, the \emph{representation similarity operator} over a dataset $X \subset \mathcal{X}$ is defined as
\begin{equation}
M_X(\theta^*, \theta) := \mathfrak{S}\Bigl(\mathbb{E}_{x \sim X}[\phi(\theta^*,x)],\, \mathbb{E}_{x \sim X}[\phi(\theta,x)]\Bigr)
\end{equation}
where $\mathfrak{S}: \mathcal{R} \times \mathcal{R} \to [0,1]$ quantifies the statistical alignment between representations and is required to satisfy the metric axioms. Similarly, the \emph{accuracy divergence operator} is formulated as
\begin{equation}
A_X(\theta^*, \theta) := \mathfrak{D}\Bigl(\mathbb{E}_{(x,y) \sim X}\ell(\psi(\theta^*,x), y),\, \mathbb{E}_{(x,y) \sim X}\ell(\psi(\theta,x), y)\Bigr)
\end{equation}
where $\mathfrak{D}: \mathbb{R}^+ \times \mathbb{R}^+ \to [0,1]$ measures the normalized task-performance discrepancy under the loss function $\ell$. The similarity space $\mathfrak{T}_X \subseteq \Theta$ is then characterized by the joint constraints
\begin{equation}
\mathfrak{T}_X(\eta, \epsilon) = \left\{ \theta \in \Theta \,\middle|\, 
\begin{aligned} 
M_X(\theta^*, \theta) &\geq 1 - \eta, \\
A_X(\theta^*, \theta) &\leq \epsilon 
\end{aligned} \right\}
\end{equation}
for a similarity tolerance $\eta \in [0,1]$ and an accuracy threshold $\epsilon \geq 0$. When considering distinct fine-tuning datasets $X_F$ with respective tolerances $(\eta_F, \epsilon_F)$, the \emph{similarity space} is defined as
\begin{align}
\mathfrak{T}_{X_F} &= \left\{ \theta \in \Theta \mid M_{X_F}(\theta^*, \theta) \geq 1 - \eta_F \,\land\, A_{X_F}(\theta^*, \theta) \leq \epsilon_F \right\}
\label{eq:simliarity_space}
\end{align}
The global similarity space is then given by Eq.~\ref{eq:simliarity_space}. This formalization establishes a theoretic structure for analyzing parameter subspaces that preserve similarity fidelity to the pretrained model while meeting task-specific performance guarantees.

\section{Experiments}
The experiments employs three vision transformer configurations and include medical dataset classification and k-NN evaluation on ImageNet-1K. We also track the variations of CKA and ECE during fine-tuning, and visualize the similarity space to quantify and predict model variations. Specifically, ViT-B/16 and OpenCLIP both process images with a resolution of $224 \times 224$ using $16 \times 16$ patch embeddings, while DINOv2-B/14 operates on inputs of size $512 \times 512$ partitioned into $14 \times 14$ patches. All models adhere to standardized training protocols with identical hyperparameters: a batch size of 32, a learning rate of $6 \times 10^{-4}$, and 20-epoch optimization via A100 GPU acceleration, implemented within unified frameworks MMClassification~\cite{Mmclassification}.

\subsection{Classification Results}
We evaluate three fine-tuning methods (Full, Linear, and Scratch) across three backbones and five datasets. To compare the contributions of different patches in the features, we examine both CLS token and patch averaging for classification. 
\begin{table*}[!ht]
    \centering

    \begin{tabular}{@{}lcccccc@{}}
        \toprule
        \textbf{Method} & \textbf{Model} & 
        \multicolumn{5}{c}{\textbf{Dataset \& Metric}} \\
        \cmidrule(lr){3-7}
        & & \textbf{ISIC2018} & \textbf{APTOS2019} & \textbf{MedFM-Chest} & \textbf{MedFM-Colon} & \textbf{MedFM-Endo} \\
        & & Acc (n=10k) & Acc (n=2.9k) & mAP (n=2.1k) & Acc (n=5.6k) & mAP (n=1.8k) \\
        \midrule
        
        \multirow{3}{*}{Full} 
        & VIT       & 83.55 (-1.56) & 83.88 (+0.16) & 26.31 (+1.00) & 99.47 (+0.20) & 57.46 (+8.82) \\
        & OPENCLIP  & 78.41 (-0.61) & 76.39 (+0.93) & 11.96 (+0.63) & 99.23 (+0.25) & 32.72 (+2.14) \\
        & DINOv2    & 75.22 (-4.04) & 74.86 (-4.10) & 12.15 (+0.67) & 95.82 (-0.88) & 20.66 (-10.02) \\
        
        \midrule
        
        \multirow{3}{*}{Linear}
        & VIT       & 73.17 (+6.61) & 77.43 (+11.20) & 22.73 (+7.42) & 93.91 (+4.40) & 37.42 (+6.61) \\
        & OPENCLIP  & 71.92 (+5.82) & 72.68 (+14.21) & 19.68 (+4.79) & 91.34 (+4.26) & 39.86 (-3.58) \\
        & DINOv2    & 73.41 (-1.56) & 78.58 (-5.68) & 20.91 (-1.21) & 93.28 (-1.82) & 33.14 (+6.72) \\
        
        \midrule
        
        Scratch & VIT & 67.83 & 72.68 & 13.64 & 91.89 & 17.67 \\
        
        \bottomrule
    \end{tabular}
    \caption{Image classification accuracy for different fine-tuning methods. Values in () indicate the improvement comparing the average of all patch features with the CLS token.}
    \label{tab:table1}
\end{table*}

\textbf{Fine-tuning and Classification Results for Medical Images.} Tab.~\ref{tab:table1} compares fine-tuning (Full, Linear) with training from scratch (Scratch). Results show Scratch consistently underperforms fine-tuning, validating the advantage of pre-trained models. Full fine-tuning outperforms Linear, reflecting differences between upstream and downstream tasks. Supervised VIT pre-training surpasses unsupervised models like DINOv2 and OPENCLIP.

For DINOv2, patch averaging outperforms the CLS token in most cases, while VIT and OPENCLIP achieve optimal results with the CLS token, highlighting the impact of pre-training.

\begin{figure*}[!t]
  \centering
  \includegraphics[width=\textwidth]{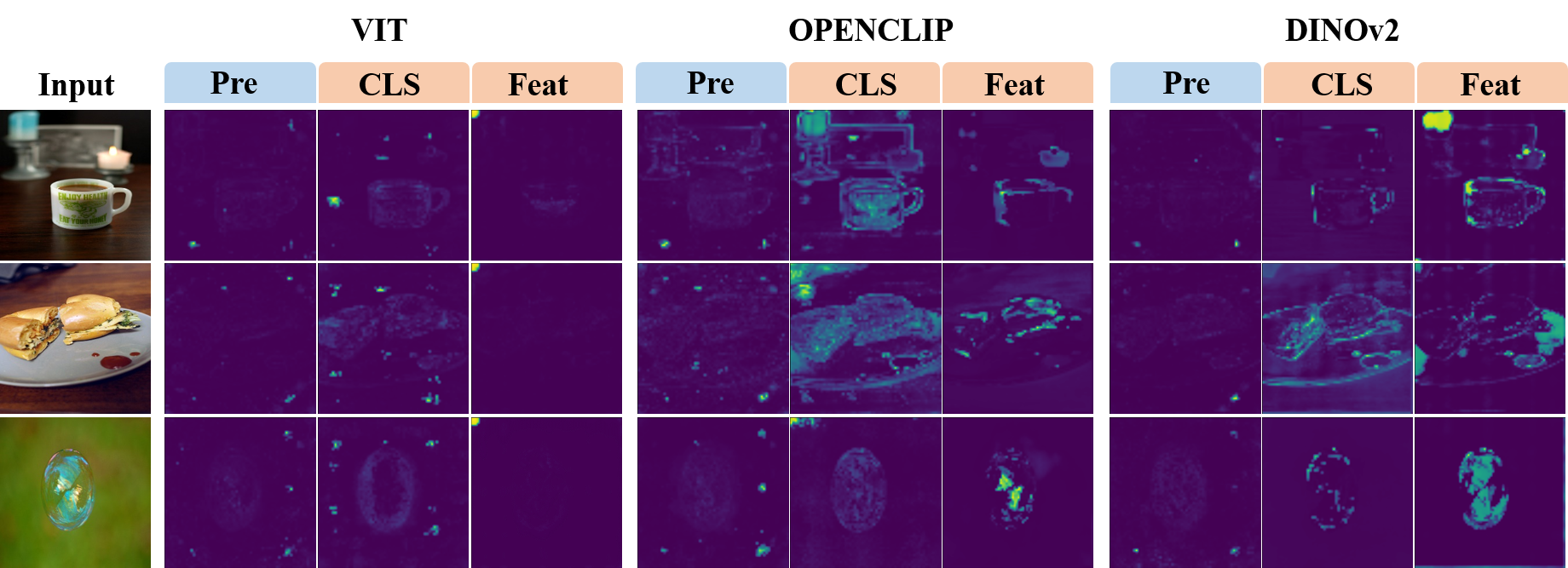}
\captionsetup{singlelinecheck=off, justification=raggedright,labelsep=period} 
  \caption{Attention maps of pre-trained and fine-tuned models are visualized using images from ImageNet-1K. Fine-tuning is performed on the ISIC2018 dataset, with CLS indicating the CLS token and Feat denoting average patch features.}
  \label{fig:fig3}
\end{figure*}

\subsection{Supervised vs. Unsupervised Pre-trained Models}
We analyze variations during fine-tuning by reporting CKA similarity, k-NN evaluation, and attention maps.

\textbf{k-NN Evaluation on ImageNet-1K.} Tab.~\ref{tab:table2} shows k-NN evaluation results on fine-tuned models. Two key findings emerge: (1) Forgetting varies slightly across datasets and models, with more complex datasets (e.g., MedFM-chest) showing greater forgetting due to larger upstream-downstream task differences. (2) Supervised pre-trained models resist forgetting better, while unsupervised models exhibit severe forgetting, reflecting differences in upstream tasks and fine-tuning behavior.

\begin{table*}[!ht]
    \centering
 {%
    \begin{tabular}{@{}lccccc@{}}
        \toprule
        \textbf{Model} &  \multicolumn{5}{c}{\textbf{Fine-tuned Datasets}} \\
        \cmidrule(lr){2-6}
                     & \textbf{ISIC2018} & \textbf{APTOS2019} & \textbf{MedFM-Chest} & \textbf{MedFM-Colon} & \textbf{MedFM-Endo} \\
        \midrule
        VIT      & 71.74 (+3.25) & 69.83 (+0.91) & 40.97 (-20.55) & 71.76 (+5.22) & 73.19 (+3.79) \\
        OPENCLIP   & 23.43 (+1.19) & 21.98 (+0.58) & 1.84 (+0.89)  & 13.81 (+2.33) & 16.63 (+0.77) \\
        DINOv2  & 5.94 (-3.08)  & 3.28 (-3.96)  & 0.65 (+0.01)  & 4.37 (-1.33)  & 3.41 (-3.51) \\
        \bottomrule
    \end{tabular}%
    }
    \caption{k-NN evaluation results of fine-tuned models on different medical datasets. Values in () indicate the improvement comparing the average of all patch features with the CLS token.}
    \label{tab:table2}
\end{table*}

\textbf{Visualization Results.}\label{sec:Visualization} Fig.~\ref{fig:fig3} visualizes attention maps before and after fine-tuning on ImageNet-1K. Pre-trained models show artifacts, focusing excessively on non-semantic regions. After fine-tuning, OPENCLIP and DINOv2 improve attention to object positions but occasionally focus on unrelated semantics (e.g., a candle instead of a coffee cup). Notably, pre-trained models perform well on ImageNet-1K classification (Tab.~\ref{tab:table2}) but generate poor visualizations, while fine-tuned models show the opposite trend.

\textbf{Differences in Behavior: Supervised or Not.} Tab.~\ref{tab:table1}, Tab.~\ref{tab:table2}, and Fig.~\ref{fig:fig3} highlight the impact of supervision. Supervised VIT achieves higher medical image classification accuracy and resists forgetting better than unsupervised OPENCLIP and DINOv2. Attention to semantic information also differs: fine-tuning improves focus, but this improvement is weaker in VIT. These differences stem from distinct pre-training objectives, leading to mismatches in downstream tasks.

\begin{figure*}[!t]
  \centering
  \includegraphics[width=1\textwidth]{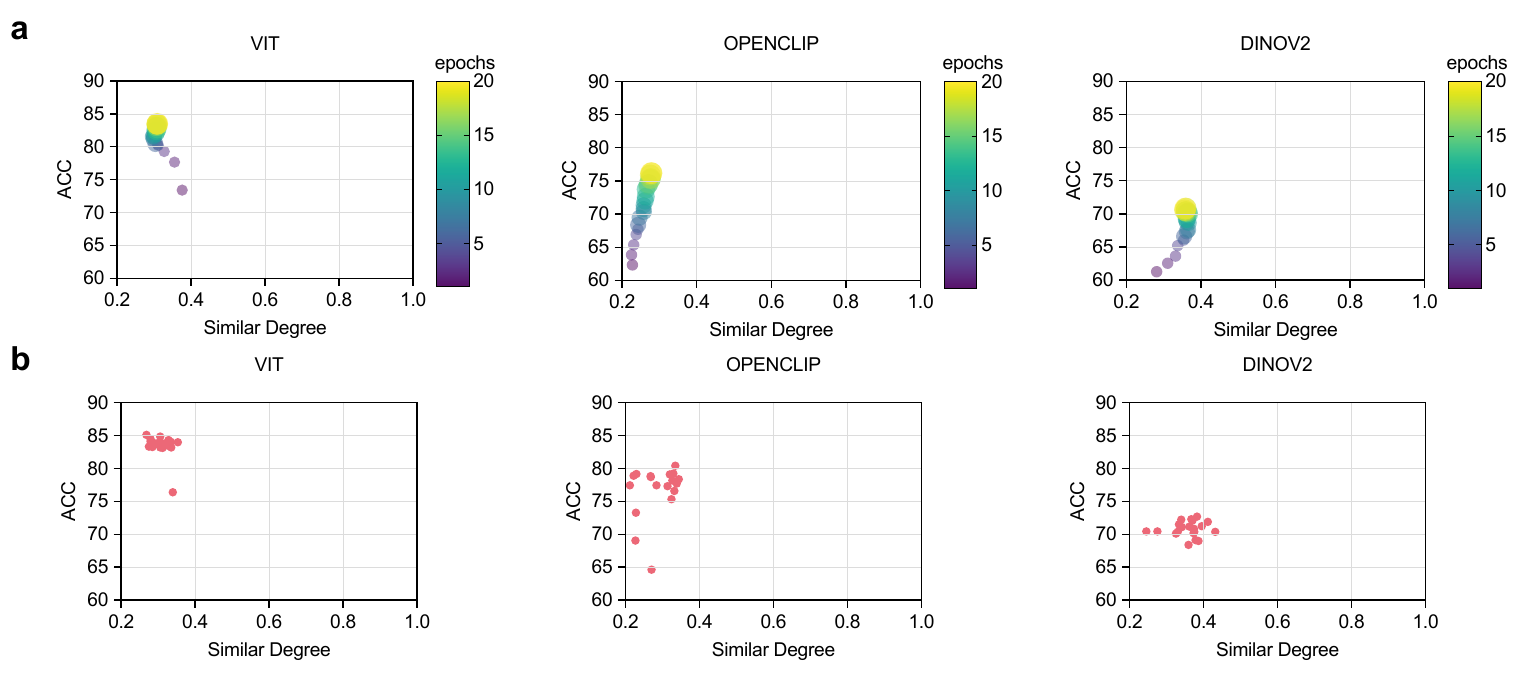}
\captionsetup{singlelinecheck=off, justification=raggedright,labelsep=period} 
  \caption{Accuracy and similarity results of fine-tuned models on ISIC2018 dataset for 20 runs. (a) The average training trajectory of 20 epochs across the 20 runs. (b) The distribution of the final similarity and accuracy of these fine-tuned models.}
  \label{fig:fig4}
\end{figure*}

\begin{figure*}[!t]
  \centering
  \includegraphics[width=1\textwidth]{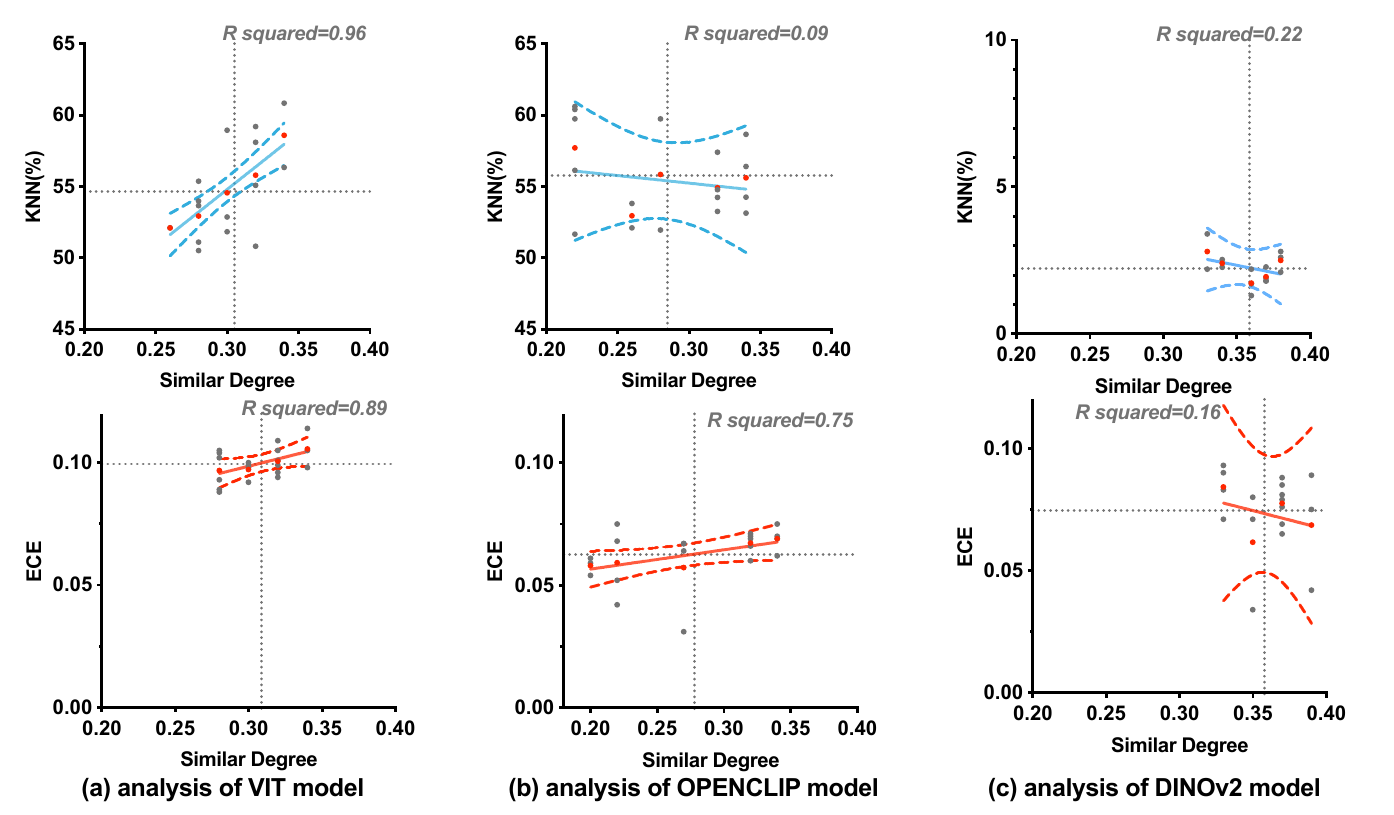}
\captionsetup{singlelinecheck=off, justification=raggedright,labelsep=period} 
  \caption{The linear correlation between similarity degree (CKA) and representation metrics (ECE) of models fine-tuned on ISIC2018 is analyzed. Similarity degrees are discretized (black points), and linear correlation is computed on interval averages (red points), with each model’s similarity assigned to the nearest interval.}
  \label{fig:fig5}
\end{figure*}
\subsection{Similarity Analysis}  
As shown in Fig.~\ref{fig:fig4} and Fig.~\ref{fig:fig5}, representation similarities are analyzed across 20 runs for each fine-tuning task, with 20 epochs per run, using the ISIC2018 dataset as an example due to its lowest similarity among all datasets.  

\textbf{Similarity Trajectory and Space.} Fig.~\ref{fig:fig4} (a) illustrates training trajectories over 20 epochs. Pre-trained models exhibit a sharp similarity decline initially, followed by gradual changes: supervised models continue decaying, while unsupervised models recover, reflecting their adaptation during supervised fine-tuning. Fig.~\ref{fig:fig4} (b) displays accuracy and similarity after fine-tuning. Pre-trained models generally show low similarity (VIT, OPENCLIP < 0.4; DINOv2 < 0.45). Models with both high accuracy and similarity, such as OPENCLIP, are identified, supporting the concept of similarity space. Future work can focus on models achieving both high similarity and accuracy.  

\textbf{Linear Correlation with Similarity.} We analyze the relationship between similarity and fine-tuning performance using k-NN evaluation and ECE results on ImageNet-1K (Fig.~\ref{fig:fig5}). Outliers are removed, and similarities are consolidated for uniformity. Linear correlations are examined for the averages (red points) of individual runs (black points). For supervised VIT, Fig.~\ref{fig:fig5}(a) illustrates that the individual similarity and k-NN evaluation scores exhibit high variability. However, their averaged values reveal a clear and robust linear relationship. A similar trend is observed for ECE. Unsupervised models exhibit varied behavior: OPENCLIP shows little correlation with k-NN but moderate correlation with ECE, while DINOv2 shows negligible correlation in both cases, highlighting distinct fine-tuning behaviors of supervised and unsupervised models.

\section{Conclusion}
This study examines similarity variations in pre-trained models during medical image fine-tuning with domain gaps. Key findings reveal that natural image pre-training significantly improves classification over training from scratch, while substantial similarity loss between fine-tuned and original models indicates underutilized pre-training potential. Our work explores behavioral differences between supervised and unsupervised models during fine-tuning and the proposed similarity space. A linear correlation between similarity degrees and specific representation metrics demonstrates predictive value.

%
%
%

%
%
%
\bibliographystyle{splncs04}
\bibliography{ref}


%




\end{document}


\title{Supplementary for\\Beyond Similarity: Exploring Fine-Tuning the Pre-trained Models for Medical Image Analysis}

\maketitle              

\begin{table}[h!]
\centering
\begin{tabular}{l|c|c|c|c|c|c}
\toprule
\textbf{Dataset} & \textbf{Modality} & \textbf{Task Type} & \textbf{Classes} & \textbf{Train} & \textbf{Test} & \textbf{Metric}\\
\hline
ISIC2018 & Dermoscopy & Multiclass & 7 & 10,015 & 1,512 & - \\
\hline
APTOS2019 & Fundus & Multiclass & 5 & 2,930 & 366 & - \\
\hline
MedFM-Chest & X-ray & Multi-label & 19 & 2,140 & 3,869 & mAP \\
\hline
MedFM-Colon & Pathology & Binary & 2 & 5,654 & - & - \\
\hline
MedFM-Endo & Endoscopy & Multi-label & 4 & 1,810 & 2,936 & mAP \\
\toprule
\end{tabular}
\caption{Extended dataset specifications.}
\label{tab:dataset_modality}
\end{table}

\begin{table}[h!]
\centering
\begin{tabular}{lccccccc}
\toprule
\textbf{Model} & \textbf{Architecture} & \textbf{Image Size} & \textbf{Patch} & \textbf{Pretrain} & \textbf{Pretrain Data} & \textbf{Batch} & \textbf{LR} \\
\midrule
ViT       & ViT-B/16 & $224^2$  & 16 & Supervised        & ImageNet-1K & 32 & $6\times10^{-4}$ \\
OpenCLIP  & ViT-B/16 & $224^2$  & 16 & Multimodal CLIP   & LAION-2B    & 32 & $6\times10^{-4}$ \\
DINOv2    & ViT-B/14 & $512^2$  & 14 & Self-distillation & LVD-142M    & 32 & $6\times10^{-4}$ \\
\bottomrule
\end{tabular}
\caption{Model configurations for comparative analysis. All vision transformers employ identical training protocols: 20 epochs on A100 with MMClassification. Architectural variants are specified through patch size (14/16) and input resolution (224/512). Pretraining strategies differ: ViT uses standard supervised learning, OpenCLIP leverages image-text contrastive alignment, and DINOv2 applies self-supervised distillation on curated data.}
\label{tab:model_config}
\end{table}

\newpage
%
%
\bibliographystyle{splncs04}
\bibliography{Paper-0151}